\pdfoutput=1

\documentclass[11pt]{article}

\usepackage[final]{acl}

\usepackage{times}
\usepackage{latexsym}

\usepackage[T1]{fontenc}

\usepackage[utf8]{inputenc}

\usepackage{microtype}

\usepackage{inconsolata}

\usepackage{graphicx}

\usepackage{pifont} 
\usepackage{amsmath} 
\usepackage{booktabs} 
\usepackage{multirow} 
\usepackage{amsfonts} 

\title{Merge then Realign: Simple and Effective Modality-Incremental Continual Learning for Multimodal LLMs}

\author{
Dingkun Zhang\textsuperscript{\rm 1},
Shuhan Qi\textsuperscript{\rm 1,\rm2}\thanks{Corresponding author.},
Xinyu Xiao\textsuperscript{\rm 1},
Kehai Chen\textsuperscript{\rm 1},
Xuan Wang\textsuperscript{\rm 1}\\
\textsuperscript{\rm 1}Harbin Institute of Technology, Shenzhen \\
\textsuperscript{\rm 2}Guangdong Provincial Key Laboratory of Novel Security Intelligence Technologies \\
\texttt{dingkunzhang0xffff@gmail.com},
\texttt{\{shuhanqi,wangxuan\}@cs.hitsz.edu.cn},\\
\texttt{23b951021@stu.hit.edu.cn},
\texttt{chenkehai@hit.edu.cn}
}

\begin{document}
\maketitle
\begin{abstract}
Recent advances in Multimodal Large Language Models (MLLMs) have enhanced their versatility as they integrate a growing number of modalities.
Considering the heavy cost of training MLLMs, it is efficient to reuse the existing ones and extend them to more modalities through Modality-incremental Continual Learning (MCL).
The exploration of MCL is in its early stages.
In this work, we dive into the causes of performance degradation in MCL.
We uncover that it suffers not only from forgetting as in traditional continual learning, but also from misalignment between the modality-agnostic and modality-specific components.
To this end, we propose an elegantly simple MCL paradigm called "MErge then ReAlign" (MERA) to address both forgetting and misalignment.
MERA avoids introducing heavy model budgets or modifying model architectures, hence is easy to deploy and highly reusable in the MLLM community.
Extensive experiments demonstrate the impressive performance of MERA, holding an average of 99.84\% Backward Relative Gain when extending to four modalities, achieving nearly lossless MCL performance.
Our findings underscore the misalignment issue in MCL.
More broadly, our work showcases how to adjust different components of MLLMs during continual learning.
\end{abstract}

\section{Introduction}
With the recent trend of developing general-purpose any-modality Multimodal Large Language Models (MLLMs) {\citep{x_instructblip, x_llm, next_gpt, one_llm, anygpt, mllm_limitation}, MLLMs are evolving towards integrating more modalities.
The typical MLLM architecture includes modality-specific encoders, modality-specific connectors, and a shared Large Language Model (LLM).
A standard process of training MLLMs involves aligning modality-specific components with LLM through modality-text paired data and then fine-tuning on modality-text instruction data \citep{commonIT}.
Such architecture and training strategy have been successfully applied to a wide range of modalities, i.e., image {\citep{llava, llava_1.5, pesf_kd}, video \citep{video_llava, video_chatgpt}, audio \citep{uni_moe, next_gpt}, point cloud \citep{damc}, etc, equipping MLLMs with the ability to understand a growing number of modalities.
Existing methods \citep{next_gpt, anygpt, x_instructblip, vita} typically employ a joint training strategy, where the MLLM is jointly trained on datasets of all predefined modalities \citep{vmt_adapter}.
However, it is challenging to extend an existing MLLM to new modalities as it requires another round of joint training on the previous modalities and the new modalities.

To reuse the existing models and adapt them to new data, Continual Learning (CL) is proposed to learn from a stream of data.
During continual learning, performance degradation in previously learned tasks often occurs.
The degradation is generally attributed to catastrophic forgetting \citep{catastrophic_forgetting_1, catastrophic_forgetting_2, apt}, i.e., the model forgets the previously learned knowledge.
To this end, many CL methods \citep{ewc, moe_adapters4cl, continual_t0, inscl} have been proposed to alleviate catastrophic forgetting.

In addition to traditional CL, Modality-incremental Continual Learning (MCL) \citep{pathweave} focuses on the particular scenario of incrementally extending MLLMs to new modalities.
The exploration of MCL is in its early stages.
In this work, we first analyze the causes of performance degradation in MCL.
Unlike traditional CL, the performance degradation encountered in MCL comes not only from forgetting but also from the misalignment between modality-agnostic and modality-specific components.

To address both forgetting and misalignment, we propose a simple yet effective two-stage MCL paradigm called "MErge then ReAlign" (MERA).

The first stage of MERA aims at addressing the forgetting problem.
To this end, we introduce model merging to our MCL framework.
We focus on the simplest model merging method, i.e., weight averaging, and revise it into an MCL form.
We achieve this by associating its merging coefficients with the progress of CL stages and only merging the modality-agnostic components.

The second stage of MERA aims at addressing the misalignment problem.
We leverage a small subset of data from each learned modality to realign the modality encoders with the LLM backbone.
In this stage, modality encoders and LLM backbone are both frozen, only the lightweight connectors are updated to enable an efficient realignment between them.
Further experiments show that the realigning stage can significantly narrow the gap between the incrementally learned MLLM and the individually trained expert MLLMs on each modality.

In summary, the contributions of this paper are threefold:
\begin{itemize}
\item We analyze the causes of degradation in Modality-incremental Continual Learning (MCL). We uncover that it suffers not only from forgetting as in traditional continual learning, but also from misalignment.
\item We propose "MErge then ReAlign" (MERA), an elegantly simple and effective two-stage MCL paradigm, to address both forgetting and misalignment.
\item Extensive experiments show that our MERA significantly outperforms other representative continual learning methods including the state-of-the-art MCL method, and \textbf{even achieves nearly lossless MCL performance}.
\end{itemize}

\begin{table*}[!t]
\centering
\begin{tabular}{lccc}
\toprule
 & Extra-Train-Memory-Free & Arch-Modification-Free & Replay-Data-Free \\ \hline
Regularization-Based & \textcolor[HTML]{ff644e}{\ding{56}} & \textcolor[HTML]{72b85d}{\ding{52}} & \textcolor[HTML]{72b85d}{\ding{52}} \\ \hline
Architecture-Based & \textcolor[HTML]{ff644e}{$\bullet$} & \textcolor[HTML]{ff644e}{\ding{56}} & \textcolor[HTML]{72b85d}{\ding{52}} \\ \hline
Replay-Based & \textcolor[HTML]{72b85d}{\ding{52}} & \textcolor[HTML]{72b85d}{\ding{52}} & \textcolor[HTML]{f5cf36}{\ding{56}} \\ \hline
Merging-Based & \textcolor[HTML]{72b85d}{\ding{52}} & \textcolor[HTML]{72b85d}{\ding{52}} & \textcolor[HTML]{72b85d}{\ding{52}} \\ \hline
MERA (Ours) & \textcolor[HTML]{72b85d}{\ding{52}} & \textcolor[HTML]{72b85d}{\ding{52}} & \textcolor[HTML]{f5cf36}{\ding{56}} \\ \bottomrule
\end{tabular}
\caption{Characteristics of different CL categories and our proposed MERA. Extra-train-memory-free: does not introduce extra GPU memory overhead at training time. Arch-modification-free: does not modify the architecture of the model or add auxiliary components. Replay-data-free: does not require access to partial data from the previous tasks or distributions. \textcolor[HTML]{ff644e}{$\bullet$} denotes that some methods of this category don't satisfy the property. \textcolor[HTML]{f5cf36}{\ding{56}} denotes that this drawback is relatively minor in real applications.}
\label{tab:comp_il}
\end{table*}

\section{Related Work}
\subsection{Multimodal Large Language Models}
Recent advances \citep{x_instructblip, x_llm, next_gpt, one_llm, anygpt} in MLLM have extended LLMs to perceive multimodal inputs such as image, video, audio, point cloud, etc.
Among all the MLLMs, the most influential one is LLaVA \citep{llava, llava_1.5}, which utilizes a simple MLP connector to project visual information encoded by the pre-trained vision encoder into the language embedding space.
Due to its simplicity and effectiveness, LLaVA-like architecture is widely adopted by a wide range of subsequent MLLMs \citep{vila, video_llava, video_chatgpt, next_gpt, damc} and accounts for the majority of current MLLMs.
In this paper, we assume that the MLLM has a LLaVA-like architecture that includes modality-specific encoders and connectors, and a shared modality-agnostic LLM backbone.

The rapid development of MLLMs demands high efficiency in their training process.
It is efficient to reuse the existing MLLMs and extend them to more modalities.
However, directly fine-tuning MLLMs on new modalities often results in significant performance degradation in previously learned modalities.
In this work, we leverage the continual learning technique to tackle this problem.

\subsection{Traditional Continual Learning}
Continual Learning (CL) \citep{three_scenarios_cl, cl_survey} aims to continually acquire new knowledge with minor forgetting of previously learned knowledge.
Existing CL methods mainly fall into the following four categories:
\textbf{Regularization-based methods} \citep{ewc, online_ewc_1, online_ewc_2} seek to protect the parameters that store important knowledge.
However, storing an importance matrix during training requires extra memory with the same scale as the trainable parameters.
\textbf{Architecture-based methods} \citep{pathweave, moe_adapters4cl, peft_moe, i2i} add task-specific parameters to the base model for each new task.
This category requires modifications to the model architecture, harming its reusability.
For many methods in this category, the model scale grows linearly as tasks increase, introducing extra memory overhead.
\textbf{Replay-based methods} \citep{replay, continual_t0, inscl} leverage a small subset of historical data and replay it when learning on new data.
This category requires access to partial data from previous tasks or distributions.
However, this drawback is relatively minor in real applications since the replay data is often accessible.
\textbf{Merging-based methods} \citep{wise_ft, magmax, lm_cocktail, model_tailor} edit models in parameter space to integrate the previously learned knowledge into the fine-tuned models by model merging \citep{ties_merging, dare_merging, adamerging, surgery_for_merging}.

Although many traditional CL methods have been proposed, they are not specifically designed for Modality-incremental Continual Learning (MCL).
In this work, we propose a simple MCL paradigm tailored for MLLMs.
Table~\ref{tab:comp_il} summarizes the characteristics of different CL categories and our proposed method.

\subsection{Continual Learning for MLLMs}
Aside from traditional continual learning, there are methods tailored for MLLMs \citep{improving_mllm_cl, modalprompt, adapt_infinity, zaf, continual_llava, pathweave}.
However, most of these works are specific to vision-language-only MLLMs, and incompatible with MCL where MLLMs can extend to arbitrarily more modalities.
To the best of our knowledge, PathWeave \citep{pathweave} is the most relevant work to MCL, and is so far the only work on MCL for MLLMs.
It is an architecture-based method that uses an adapter-in-adapter mechanism to alleviate forgetting in previous modalities.

In this work, we dive into the causes of degradation in MCL, uncovering that it suffers not only from forgetting but also from misalignment.
To this end, we propose our two-stage MCL paradigm to address both forgetting and misalignment.

\begin{figure*}[!t]
\centering
\includegraphics[width=\linewidth]{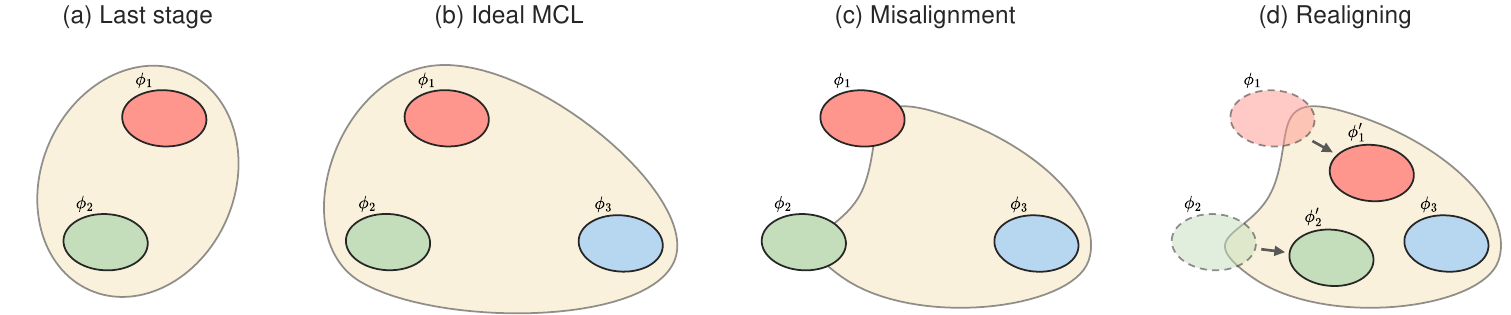}
\caption{Illustration of misalignment and the mechanism of our proposed realigning. $\phi_i$ is the feature distribution of the $i$-th modality. Regions in \textcolor[HTML]{e3c053}{yellow} represent the LLM's expected distribution of the connector's output. (a) and (b) are the states of the last learning stage and the ideal MCL after learning on a new modality. (c) shows the actual misalignment after learning on a new modality. (d) demonstrates the mechanism of our proposed realigning.}
\label{fig:misalignment}
\end{figure*}

\section{Dual Causes of Degradation in MCL}

\subsection{Preliminary}
We define the Modality-incremental Continual Learning (MCL) problem as follows.
Given a sequence of $m$ modalities $\{M_1, M_2, \dots, M_m\}$ and their corresponding datasets $\{D_1, D_2, \dots, D_m\}$, MCL sequentially learns on each modality $M_i$ to obtain the model $\theta_i$.
We denote the MLLM as $\theta=\{\theta^{Enc},\theta^{Conn},\theta^{LLM}\}$, where $\theta^{Enc},\theta^{Conn},\theta^{LLM}$ denote the modality encoders, the connectors, the LLM backbone, respectively.
Notably, the $\theta^{Enc}$ and $\theta^{Conn}$ are modality-specific components and the $\theta^{LLM}$ is a modality-agnostic component.
Further, we denote the feature distribution of the outputs of each modality connector $\theta_i^{Conn}$ as $\phi_i$.

\subsection{Forgetting and Misalignment in MCL}
MCL is a special scenario of CL, where models incrementally learn on new modalities.
However, MCL faces a more severe problem.
During continual learning, models would suffer from performance degradation in previously learned domains, tasks, or modalities.
In traditional CL, degradation comes from the forgetting of previously learned knowledge.
However, in MCL, it comes from two significant aspects: forgetting and misalignment.

\textbf{Forgetting}:
the modality-agnostic $\theta^{LLM}$ forgets the knowledge of old modalities.
It is associated with various factors, including representation drift \citep{representation_change_in_cl}, gradient interference \citep{olora}, learning dynamics \citep{learning_dynamics}, distribution shift, etc.

\textbf{Misalignment}:
the modality-agnostic $\theta^{LLM}$ is misaligned with the modality-specific $\theta^{Enc}$.
When incrementally learning on a new modality $M_i$, the $\theta^{LLM}$ is updated along with the modality connector of $M_i$, while other old connectors are kept frozen.
Therefore, $\theta^{LLM}$ adapts to the new feature distribution $\phi_i$ and drifts away from the original multimodal distribution $\phi_1\cup \phi_2\cup \dots\cup \phi_{i-1}$\footnote{For our proposed method, the $\theta^{LLM}$ also drifts away from $\phi_i$ due to the use of model merging.}.
Hence, there exists a misalignment in feature mapping between the old modality encoders and the $\theta^{LLM}$, leading to the breakdown of the "encoder-connector-LLM" collaboration chains for old modalities.
The illustration of misalignment is sketched in Figure~\ref{fig:misalignment}.
Experiments in Section~\ref{sec:misalignment_is_common} also provide empirical evidence of the existence of misalignment.

Above, we analyzed the dual causes of performance degradation in MCL.
Due to the existence of misalignment, MCL problem requires special treatments in contrast to traditional CL problems.

\section{Method}
To tackle the dual causes of performance degradation in MCL, we propose a two-stage MCL paradigm called "MErge then ReAlign" (MERA).
In each stage of MCL, MERA executes the following two stages: merging and realigning, to address forgetting and misalignment respectively.

\subsection{Stage 1: Merging}
Model merging is efficient in integrating the previously learned knowledge into the fine-tuned models.
Moreover, \citet{merging_at_scale} finds that model merging is more effective with larger models.
Therefore, applying model merging to large-scale models such as MLLMs is inherently beneficial.
Inspired by these, we introduce model merging to our MCL framework to mitigate forgetting.

The first step before model merging is to perform the standard MLLM training, which often encompasses a pre-training and a fine-tuning phase.
After the standard training step, we get the vanilla model $\theta_{i, vanilla}$ that inevitably suffers from forgetting the knowledge of previous modalities.

The second step is to perform model merging to mitigate forgetting.
In this work, we only focus on the simplest model merging method, i.e., weight averaging, \textit{aiming only to provide a basic framework}.
To adapt weight averaging to MCL, we associate its merging coefficients with the progress of MCL stages.
At the $i$-th training stage, the merged model is calculated by:
\begin{equation*}
\theta_{i, merged} = \frac{i-1}{i}\theta_{i-1} + \frac{1}{i}\theta_{i, vanilla}
\end{equation*}
Notably, we only merge the modality-agnostic component $\theta^{LLM}$ and ensemble the modality-specific components $\theta^{Enc}$ and $\theta^{Conn}$.
After merging, we obtain the $\theta_{i, merged}$, whose knowledge of previous modalities is enhanced.

\subsection{Stage 2: Realigning}
To address the misalignment issue, we propose a lightweight realigning stage to update the multimodal distribution $\phi_1\cup \phi_2\cup \dots\cup \phi_i$ to realign the "encoder-connector-LLM" chains for all the modalities.

The realigning stage simply leverages a small replay dataset\footnote{It is different from replay data in replay-based continual learning methods, where their $R_i$ is sampled from $\{D_1, D_2, \dots, D_{i-1}\}$.} $R_i\leftarrow$ sample $r$\% data from $\{D_1, D_2, \dots, D_i\}$ to further fine-tune all the connectors of $\theta_{i, merged}$.
This realigning process is formulated as:
\begin{equation*}
\min\limits_{\theta_{i, merged}^{Conn}} \mathbb{E}_{x\sim R_i} \mathcal{L}(\theta_{i, merged}, x)
\end{equation*}
where the $\mathcal{L}$ is the auto-regressive loss, unchanged from the original MLLM training loss.
By fine-tuning the lightweight $\theta^{Conn}$ with only a small replay dataset, it efficiently realigns the $\theta^{Enc}$ with $\theta^{LLM}$.
After realigning, we can obtain the final model $\theta_i$.

\begin{figure*}[!t]
\centering
\includegraphics[width=\linewidth]{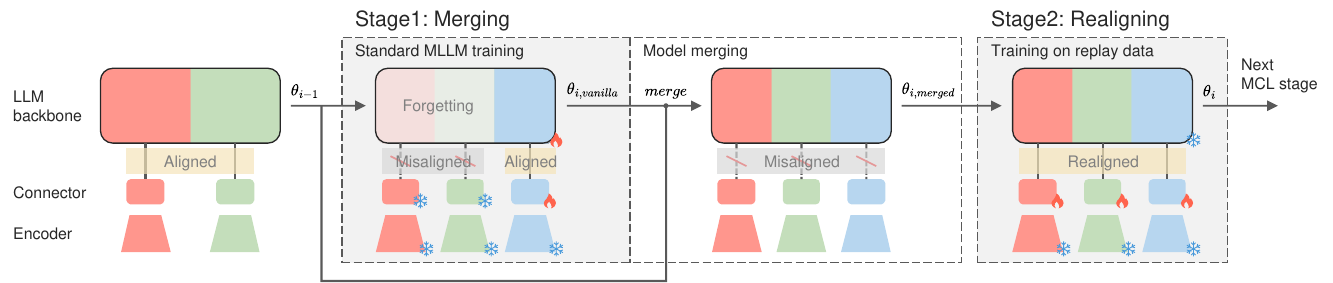}
\caption{Pipeline of the proposed MERA. The procedures in \textcolor{gray}{gray} boxes involve training. \includegraphics[height=\ht\strutbox]{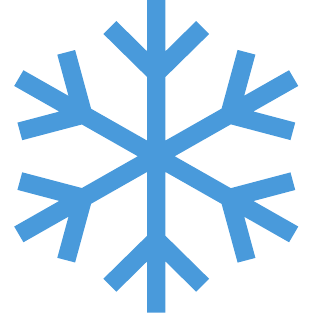} and \includegraphics[height=\ht\strutbox]{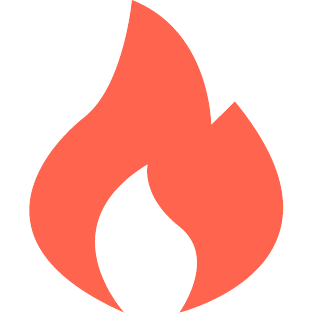} represent the frozen and trainable modules, respectively.}
\label{fig:method}
\end{figure*}

\textbf{Differences between realigning and replay-based CL methods.}
The realigning stage resembles replay-based CL methods \citep{continual_t0, inscl} in form as they both leverage replay data, however, they are essentially different. First, replay methods train on the joint dataset of $D_i$ and $R_i$, while our realigning stage trains solely on $R_i$.
Second, replay methods update both the $\theta^{LLM}$ and $\theta^{Conn}$, while the realigning stage only efficiently updates the $\theta^{Conn}$ to prevent the knowledge inside $\theta^{LLM}$ from being overwritten again.

\subsection{Overall Pipeline}
The overall pipeline of MERA is illustrated in Figure~\ref{fig:method}.
In each stage of MCL, MERA goes through two stages.
The first stage is merging, where we fine-tune $\theta_{i-1}$ on the incoming modality and merge the fine-tuned model $\theta_{i-1, vanilla}$ with the historical model $\theta_{i-1}$ in order to alleviate forgetting.
The second stage is realigning, where we leverage a small set of replay data $R_i$ to efficiently fine-tune the lightweight modality connectors $\theta^{Conn}$ to realign the $\theta^{Enc}$ with $\theta^{LLM}$.

\section{Experiments}
\subsection{Experimental Setup}
We build our MCL experiments on four modalities: image, video, audio, and point cloud, with two different training orders.
Based on the prevalence of different modalities, we determine the two orders as follows.
\textbf{Sequential Order}: image $\rightarrow$ video $\rightarrow$ audio $\rightarrow$ point cloud.
\textbf{Reverse Order}: point cloud $\rightarrow$ audio $\rightarrow$ video $\rightarrow$ image.
On top of this, the adopted datasets, metrics, models, and baselines are detailed as follows.

\textbf{Datasets.}
For each modality $M_j$, we leverage a dataset of Captioning (Cap) task and a dataset of Question Answering (QA) task to form the joint dataset $D_j=\{D_{j,Cap}, D_{j,QA}\}$.
The Cap and QA datasets for each modality are listed respectively.
For image modality, we use MSCOCO-2014 \citep{mscoco} and OK-VQA \citep{ok_vqa}.
For video modality, we use MSVD \citep{msvd} and MSVD-QA \citep{msvd_qa}.
For audio modality, we use AudioCaps \citep{audiocaps} and Clotho-AQA \citep{clotho_aqa}.
For point cloud modality, we use a subset of Cap3D \citep{cap3d} and a subset of Cap3D-QA \citep{x_instructblip}.
More details of these datasets are in Appendix~\ref{appendix:datasets}.

\textbf{Evaluation Metrics.}
First, we leverage Relative Gain \citep{continual_t0, inscl} as a normalized metric across different tasks.
We naively train (without using any continual learning methods) expert MLLMs individually on each single modality $M_j$ and test with their respective holdout data, taking their scores on the $k$-th dataset $D_{j,k}$ as upper bound $S_{j,k}^{sup}$.
In the incremental stage $i$, the Relative Gain of modality $M_i$ with its dataset $D_j=\{D_{j,k}\}_{k=1}^K$ is calculated by:
\begin{equation*}
\mathrm{Relative\ Gain}_j^i = \frac{1}{K}\sum_{k=1}^{K}\frac{S_{j,k}^i}{S_{j,k}^{sup}}
\end{equation*}
where $S_{j,k}^i$ is the score on the test set of $D_{j,k}$ in the stage $i$.
Here, we utilize CIDEr score \citep{cider} and prediction accuracy (Acc) for Cap and QA tasks respectively to calculate $S_{j,k}^{sup}$ and $S_{j,k}^i$.
To evaluate the performance degradation of the previously learned modalities, we calculate the Backward Relative Gain in the stage $i$ as:
\begin{equation*}
\mathrm{Bw\ Relative\ Gain}^i = \frac{1}{i-1}\sum_{j=1}^{i-1}\mathrm{Relative\ Gain}_j^i
\end{equation*}
To measure the plasticity, i.e., the ability to adapt to new knowledge, we calculate the Forward Relative Gain in the stage $i$ as:
\begin{equation*}
\mathrm{Fw\ Relative\ Gain}^i = \mathrm{Relative\ Gain}_i^i
\end{equation*}

\begin{figure*}[!t]
\centering
\includegraphics[width=\linewidth]{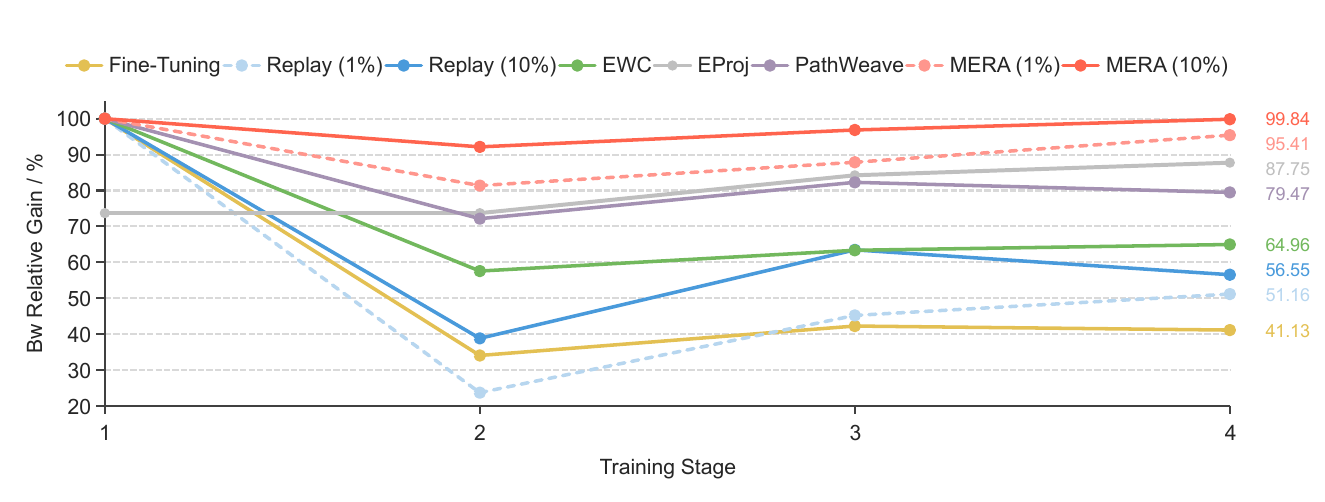}
\caption{Progressive Backward Relative Gain in modality-incremental continual learning. For each stage $i$, we plot the average score of the corresponding Backward Relative Gain with two different training orders. We set Backward Relative Gain to 100\% for the $1$st stage, denoting the initial performance without degradation. Exceptionally, the initial Backward Relative Gain of EProj is not 100\% since it only tunes the modality-specific components, causing an initial performance degradation.}
\label{fig:bw_relative_gain}
\end{figure*}

\textbf{Model and Training Details.}
We leverage the mainstream MLLM architecture, i.e., LLaVA-like architecture with the Llama-3-8B-Instruct \citep{llama3} as its LLM backbone.
The selections of modality encoders and connectors are detailed in Appendix~\ref{appendix:implementation}.
Trainings that involve updating the LLM backbone utilize LoRA \citep{lora} for parameter-efficient fine-tuning.
The training process in our merging stage is the same as the standard MLLM training, i.e., in the first step, only the connector is updated with Cap datasets, then in the second step, the connector and the LLM backbone are updated with all the task-related datasets (the combination of Cap and QA datasets in our case).
In our realigning stage, the replay datasets are randomly sampled from the joint datasets of Cap and QA tasks.
For each training process, the hyperparameters are listed in Appendix~\ref{appendix:implementation}.

\textbf{Baselines.}
In our experiments, we compare our MERA with non-CL fine-tuning, as well as the representative CL and MCL methods:
\textbf{Fine-Tuning}: directly train MLLMs sequentially on each modality without applying any CL method.
\textbf{Replay}: the vanilla replay-based CL method. During training on a new task, the model is updated with both samples from the current task and a set of randomly sampled replay data from previous tasks.
\textbf{EWC} \citep{ewc}: the most representative regularization-based CL method. EWC mitigates forgetting by restricting the updates of important weights during training on new tasks. It uses the Fisher information matrix to measure the importance of each weight.
\textbf{EProj} \citep{eproj}: tuning only the modality-specific components to prevent forgetting.
\textbf{PathWeave} \citep{pathweave}: an architecture-based CL method, also the state-of-the-art MCL method for MLLMs. PathWeave uses an adapter-in-adapter mechanism to memorize and extract knowledge from historical modalities to enhance the learning of the current modality. PathWeave is originally built on X-InstructBLIP \citep{x_instructblip}. For a fair comparison, we implement PathWeave for our adopted MLLM architecture.
Implementation details of each baseline method are in Appendix~\ref{appendix:baselines}.

\begin{figure*}[!t]
\centering
\includegraphics[width=\linewidth]{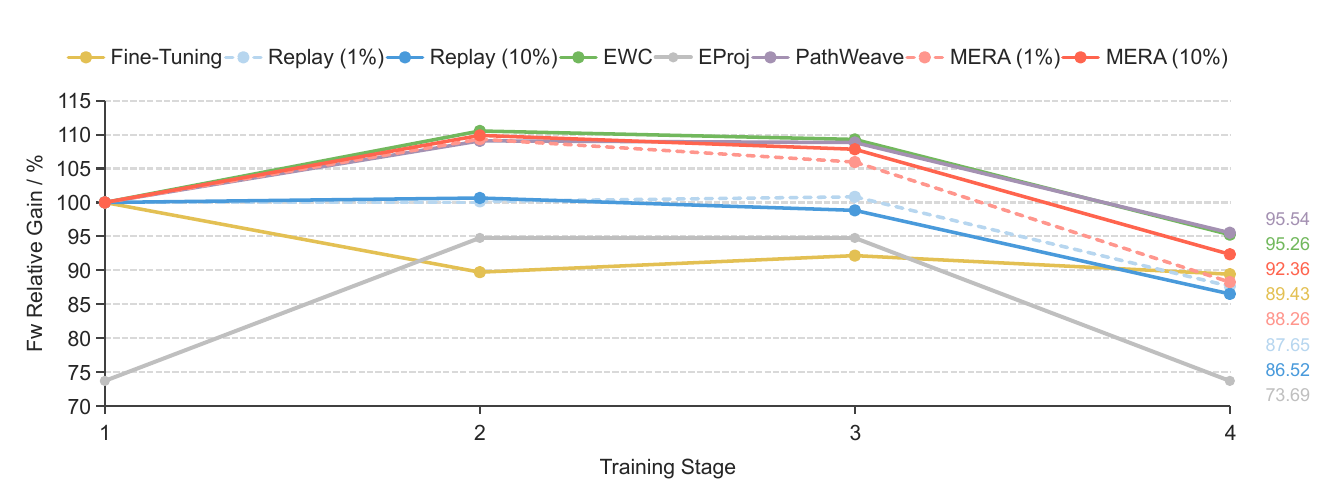}
\caption{Progressive Forward Relative Gain in modality-incremental continual learning. For each stage $i$, we plot the average score of the corresponding Forward Relative Gain with two different training orders. We set Forward Relative Gain to 100\% for the $1$st stage, denoting the initial lossless plasticity. Exceptionally, the initial Forward Relative Gain of EProj is not 100\% since it only tunes the modality-specific components, causing an initial loss of plasticity.}
\label{fig:fw_relative_gain}
\end{figure*}

\subsection{Main Results}
We conduct experiments under our MCL setting with both sequential and reverse orders.
For Replay and our MERA, results using $r$\%, $r=\{1,10\}$ replay data are reported, denoted by Replay ($r$\%) and MERA ($r$\%) respectively.

\textbf{Evaluating Degradation.}
The progressive Backward Relative Gains averaged from different training orders are plotted in Figure~\ref{fig:bw_relative_gain}.
It is observed that our MERA demonstrates an impressive capability of mitigating performance degradation with consistent and promising Backward Relative Gains.
When extending to all four modalities, MERA ($10\%$) holds up to a 99.84\% Backward Relative Gain, indicating that MERA can achieve nearly lossless MCL performance, with at least 12.09\% absolute improvements of Backward Relative Gain compared with other baselines.
Notably, when only leveraging $1\%$ replay data, MERA (1\%) can still achieve at least 7.66\% absolute improvements over other baselines.
Further, we calculated the mean and standard deviation of Backward Relative Gains in all training stages for each method, in different training orders.
Table~\ref{tab:bw_relative_gain} shows that our MERA (10\%) achieves the highest mean in both training orders, indicating its superior performance.
It also achieves the lowest and second-lowest standard deviation in sequential and reverse orders respectively, indicating its high stability.
Notably, in sequential order, MERA (10\%) performs even better than lossless MCL, with an over 100\% average Backward Relative Gain, also at least 11.39\% absolute improvements of average Backward Relative Gain over other baselines.
In reverse order, MERA (10\%) also achieves at least 13.33\% absolute improvements.
When with only 1\% replay data, MERA (1\%) still achieves at least 8.29\% and 4.33\% absolute improvements in sequential and reverse orders respectively.

\begin{table}[!t]
\centering
\resizebox{\columnwidth}{!}{
\begin{tabular}{lcccc}
\toprule
\multirow{2}{*}{\textbf{Method}} & \multicolumn{2}{c}{\textbf{Sequential}} & \multicolumn{2}{c}{\textbf{Reverse}} \\ \cline{2-5} 
 & \textbf{Mean} & \textbf{Std} & \textbf{Mean} & \textbf{Std} \\ \midrule
Fine-Tuning & 59.76 & 27.23 & 48.96 & 35.70 \\
Replay (1\%) & 66.09 & 25.86 & 43.95 & 39.03 \\
Replay (10\%) & 77.52 & 16.32 & 51.90 & 36.34 \\
EWC & 74.93 & 17.14 & 68.01 & 21.54 \\
EProj & 89.61 & \textbf{2.63} & 70.07 & \underline{11.86} \\
PathWeave & 86.85 & 12.17 & 80.09 & 13.31 \\ \midrule
MERA (1\%) & \underline{97.90} & 6.02 & \underline{84.42} & 12.93 \\
MERA (10\%) & \textbf{101.00} & \underline{3.90} & \textbf{93.42} & \textbf{6.25} \\ \bottomrule
\end{tabular}
}
\caption{The mean and standard deviation of Backward Relative Gains in all the training stages. Results are reported on different training orders. The best results are in \textbf{bold}, while the second-best are \underline{underlined}.}
\label{tab:bw_relative_gain}
\end{table}

\begin{table}[!t]
\centering
\begin{tabular}{lcc}
\toprule
\textbf{Method} & \textbf{Sequential} & \textbf{Reverse} \\ \midrule
Fine-Tuning & 38.50 & 36.14 \\
Replay (1\%) & 30.61 & 33.02 \\
Replay (10\%) & 56.68 & 32.50 \\
EWC & 25.79 & 24.94 \\
EProj & 55.61 & \underline{55.61} \\
PathWeave & 36.32 & 46.39 \\ \midrule
MERA (1\%) & \underline{72.13} & \textbf{61.22} \\
MERA (10\%) & \textbf{72.70} & 54.04 \\ \bottomrule
\end{tabular}
\caption{Accuracies on MCUB-4 benchmark. Results are reported on different training orders and are measured after continually learning on all four modalities.}
\label{tab:multimodal_task}
\end{table}

\textbf{Evaluating Degradation on Complex Multimodal Tasks.}
To evaluate MERA's performance in more complex multimodal tasks that involve multiple modalities at a time, we further conduct experiments on the MCUB-4 \citep{damc} benchmark that requires the model to simultaneously infer on all four modalities, i.e., image, video, audio, and pointcloud.
Table~\ref{tab:multimodal_task} shows the accuracies of each method on MCUB-4 benchmark.
It is observed that our proposed MERA also significantly excels in more complex multimodal tasks that involve multiple modalities at a time.

\textbf{Evaluating Plasticity.}
Aside from alleviating performance degradation, the capability to adapt to new knowledge, i.e., plasticity, is also an important aspect.
We use the Forward Relative Gain as the metric.
The progressive Forward Relative Gains averaged from different training orders are plotted in Figure~\ref{fig:fw_relative_gain}.
It is observed that the most elastic CL methods are EWC and PathWeave, while our MERA (10\%) demonstrates comparable plasticity.

\subsection{Ablation Study}
We conduct ablation studies to investigate the effectiveness of each stage of MERA.
Results are shown in Table~\ref{tab:ablation}.
Firstly, from Table~\ref{tab:ablation} and Table~\ref{tab:bw_relative_gain}, it is observed that the realigning stage that addresses the misalignment issue can already beat many other baselines designed to tackle the forgetting issue, achieving the second-best average Backward Relative Gain among baselines in both sequential and reverse orders.
Secondly, combining both merging and realigning stages, MERA further narrows the gap between the incrementally learned models and the individually trained experts on each modality, even, surpassing the individually trained experts in sequential training order with over 100\% Backward Relative Gain.

\begin{table}[!t]
\centering
\resizebox{\columnwidth}{!}{
\begin{tabular}{lcccc}
\toprule
\multirow{2}{*}{\textbf{Method}} & \multicolumn{2}{c}{\textbf{Sequential}} & \multicolumn{2}{c}{\textbf{Reverse}} \\ \cmidrule(lr){2-3} \cmidrule(lr){4-5}
 & \textbf{Mean} & \textbf{Std} & \textbf{Mean} & \textbf{Std} \\ \midrule
Fine-Tuning & 59.76 & 27.23 & 48.96 & 35.70 \\ \midrule
+Merging & 90.29 & 7.42 & 70.00 & 30.87 \\
+Realigning & 87.92 & 12.41 & 71.90 & 24.52 \\ \midrule
MERA & \textbf{101.00} & \textbf{3.90} & \textbf{93.42} & \textbf{6.25} \\ \bottomrule
\end{tabular}
}
\caption{Ablation study of different components in MERA. The realigning stage uses 10\% replay data.}
\label{tab:ablation}
\end{table}

\section{Discussions}

\subsection{Is Misalignment Common in MCL?}\label{sec:misalignment_is_common}
Since the realigning stage achieves great success on top of our proposed merging stage, we further ask another question: does realigning benefit other CL methods, or in other words, \textit{is misalignment a common phenomenon in MCL}?
To examine this, we perform the realigning stage at the end of every training stage for different CL or non-CL methods to observe whether there are performance improvements\footnote{We do not examine this for EProj. EProj is self-evidently misalignment-free, since it freezes the LLM backbone.}.
Table~\ref{tab:misalignment} shows that the additional realigning stage brings substantial performance improvements and increased stability for different CL or non-CL methods.
Based on this observation, we can conclude that \textbf{misalignment is a common phenomenon in MCL}, and can be alleviated by our proposed realigning stage.

\begin{table}[!t]
\centering
\resizebox{\columnwidth}{!}{
\begin{tabular}{lcccc}
\toprule
\multirow{2}{*}{\textbf{Method}} & \multicolumn{2}{c}{\textbf{Sequential}} & \multicolumn{2}{c}{\textbf{Reverse}} \\ \cmidrule(lr){2-3} \cmidrule(lr){4-5}
 & \textbf{Mean} & \textbf{Std} & \textbf{Mean} & \textbf{Std} \\ \midrule
Fine-Tuning & 59.76 & 27.23 & 48.96 & 35.70 \\
+Realigning & \textcolor[HTML]{72b85d}{+28.16} & \textcolor[HTML]{72b85d}{-14.83} & \textcolor[HTML]{72b85d}{+22.93} & \textcolor[HTML]{72b85d}{-11.17} \\ \midrule
Replay (1\%) & 66.09 & 25.86 & 43.95 & 39.03 \\
+Realigning & \textcolor[HTML]{72b85d}{+20.64} & \textcolor[HTML]{72b85d}{-9.75} & \textcolor[HTML]{72b85d}{+24.35} & \textcolor[HTML]{72b85d}{-8.36} \\ \midrule
Replay (10\%) & 77.52 & 16.32 & 51.90 & 36.34 \\
+Realigning & \textcolor[HTML]{72b85d}{+14.21} & \textcolor[HTML]{72b85d}{-6.67} & \textcolor[HTML]{72b85d}{+23.71} & \textcolor[HTML]{72b85d}{-11.07} \\ \midrule
EWC & 74.93 & 17.14 & 68.01 & 21.54 \\
+Realigning & \textcolor[HTML]{72b85d}{+19.54} & \textcolor[HTML]{72b85d}{-6.87} & \textcolor[HTML]{72b85d}{+23.02} & \textcolor[HTML]{72b85d}{-13.25} \\ \midrule
PathWeave & 86.85 & 12.17 & 80.09 & 13.31 \\
+Realigning & \textcolor[HTML]{72b85d}{+6.22} & \textcolor[HTML]{72b85d}{-0.28} & \textcolor[HTML]{72b85d}{+2.40} & \textcolor[HTML]{72b85d}{-1.53} \\ \midrule
Merging & 90.29 & 7.42 & 70.00 & 30.87 \\
+Realigning & \textcolor[HTML]{72b85d}{+10.71} & \textcolor[HTML]{72b85d}{-3.52} & \textcolor[HTML]{72b85d}{+23.42} & \textcolor[HTML]{72b85d}{-24.63} \\ \bottomrule
\end{tabular}
}
\caption{Applying realigning to different CL methods can further improve their Backward Relative Gain and stability. The realigning stage uses 10\% replay data. Improvements are colored in \textcolor[HTML]{72b85d}{green}.}
\label{tab:misalignment}
\end{table}

\subsection{Positive Backward Transfer and Positive Forward Transfer}\label{sec:positive_transfer}
From Figure~\ref{fig:bw_relative_gain} and its raw data shown in Appendix~\ref{appendix:raw_data}, we observe a faint phenomenon of Positive Backward Transfer \citep{backward_transfer} that learning new knowledge improves the performance on previously learned tasks.
For most methods, the Backward Relative Gain comes to a low level when incrementally learning the second modality, but starts to stabilize and even increase when incrementally learning more modalities.

From Figure~\ref{fig:fw_relative_gain}, we observe a strong phenomenon of Positive Forward Transfer \citep{forward_transfer} that the knowledge acquired from earlier tasks improves the learning efficiency of new tasks.
The Positive Forward Transfer exists before the $4$-th incremental stage, when employing EWC, PathWeave, and MERA.
This phenomenon is also reported in other MCL literature \citep{pathweave}.
In contrast to Positive Forward Transfer, there is a gradual loss of plasticity \citep{plasticity_loss_1, plasticity_loss_2} as the model attempts to retain more knowledge.
This explains the decreases in Forward Relative Gain across different CL methods in the $4$-th stage, as the loss of plasticity comes to a dominant position.

\section{Future Work}
This work is one of the early attempts of MCL for MLLMs, focusing on the dual causes of its degradation.
In addition to this, observation of Positive Backward Transfer and Positive Forward Transfer in Section~\ref{sec:positive_transfer} may imply the complex cross-modal interaction in multimodal learning, urging for future research on the mechanisms of modality interaction in the context of MCL.

\section{Conclusion}
In this paper, we first revisit MCL and uncover the dual causes of its degradation, i.e., forgetting and misalignment.
Next, to address both forgetting and misalignment, we propose MERA, a simple yet effective MCL paradigm.
Extensive experiments demonstrate that MERA significantly outperforms the state-of-the-art methods and even achieves nearly lossless MCL performance.
Our findings underscore the misalignment issue in MCL.
More broadly, our work showcases how to adjust different components of MLLMs during continual learning.
Further, we observe signs of complex cross-modal interaction in MCL, providing a direction for future work.

\section{Limitations}
This work is restricted in the following aspects.
First, our experiments are limited to four commonly used modalities due to the lack of resources for other less-studied modalities.
Second, we limit this work to LLaVA-like architecture as it covers the majority of MLLMs.
Third, this work is limited to any-to-text MLLMs while there is now a trend of exploring any-to-any MLLMs.
However, the main idea of MERA is generic to them.
Fourth, our work only explores the simplest model merging method in the context of MCL, aiming to provide a universal framework, leaving other model merging methods for MCL for future work.

\section{Acknowledgments}
This work was supported by the National Natural Science Foundation of China (No.62372139), the National Natural Science Foundation of China (2024A1515030024), Research Projects of Shenzhen (JCYJ20220818102414030), and Key Laboratory of Guangdong Province (2022B1212010005).


\clearpage
\appendix
\section*{Appendix}

\section{Further Analysis and Discussions}

\subsection{Deeper Discussions on the Distinction Between Forgetting and Misalignment}
\textbf{Theoretically}, forgetting is associated with various factors such as representation drift \citep{representation_change_in_cl}, gradient interference \citep{olora}, learning dynamics \citep{learning_dynamics}, distribution shift (which causes misalignment), making it a comprehensive issue.
For example, \citet{multimodal_neurons} finds that there exist multimodal neurons in the LLM backbone, each corresponds to certain domain concepts, and when breaking some of these neurons, the model's prediction is largely affected.
However, addressing the misalignment issue by our proposed realigning will possibly not recover the model's performance since the connectors do not encode domain-specific concepts \citep{projection_in_mllm}.
This example showcases that the forgetting issue is not limited to misalignment.
\textbf{Empirically}, our experiments in Table~\ref{tab:misalignment} have shown that, a) addressing only the misalignment issue, i.e., fine-tuning with realigning, does not outperform some methods that address the overall forgetting, e.g., PathWeave/merging without realigning, b) methods that address the overall forgetting significantly complement the realigning alone (fine-tuning with realigning).
In terms of methodology, forgetting is a comprehensive issue that can be tackled mainly by heuristic methods such as EWC or replay, while misalignment is caused solely by distribution shift that can be directly addressed by realigning the LLM and modality-specific components.
\textbf{In terms of methodology}, forgetting is a comprehensive issue that can be tackled mainly by heuristic methods such as EWC or replay, while misalignment is caused solely by distribution shift that can be directly addressed by realigning the LLM and modality-specific components.

\subsection{Misalignment Also Exists In Replay-Based Methods}
Theoretically, a potential exception to misalignment is when replay-based methods are applied, since the $\theta^{LLM}$ is trained on the joint distribution of all $\{M_1, M_2, \dots, M_i\}$ modalities.
However, Table~\ref{tab:misalignment} suggests that replay-based methods still suffer from misalignment and can be compensated for by our proposed realigning stage.
We conjecture that it is due to the imbalanced distribution of its training data.
Replay-based methods train the model on the joint dataset of its replay data and $D_i$, where the scale of replay data from each previous modality is insignificant to the scale of $D_i$.
Therefore, it still suffers from a certain degree of distribution shift during its training process.

\subsection{Easily Adapt MERA to Other MLLM Architectures}
Although we limit our work to LLaVA-like architecture, our method can be easily adapted to other MLLM architectures.
We present several naive ways to adapt MERA to some other MLLM architectures.
\begin{itemize}
\item For connector-free MLLMs, the last few layers of their encoders can be treated as connectors so that our method can be directly applied.
\item For MLLMs that use a uni-connector for all modalities, we can treat them as the connector-free MLLMs.
\item For encoder-free MLLMs, we can realign the raw multimodal input distributions instead of multimodal feature distributions with the LLM backbone by fine-tuning the embedding layers.
\end{itemize}

\subsection{Efficiency Comparisons}
We compare the efficiency of different baselines and our MERA, as shown in Table~\ref{tab:efficiency_comparison}.
It is observed that our MERA can achieve optimal results except that MERA (1\%) and MERA (10\%) introduce 2\% and 15\% extra training time-consuming respectively.
However, we believe its trade-off between training time-consuming and performance is worthwhile, considering the impressive performance of MERA.
In our experiments, EWC and PathWeave introduce marginal extra training memory overhead, as we employ parameter-efficient fine-tuning.
However, for larger LoRA ranks or even full model fine-tuning, their extra training memory consumptions would be substantial, as they necessitate storing additional parameters whose sizes increase linearly with the trainable parameters.

\begin{table*}[!t]
\centering
\begin{tabular}{lcccc}
\toprule
\multirow{2}{*}{\textbf{Method}} & \multicolumn{2}{c}{\textbf{Training}} & \multicolumn{2}{c}{\textbf{Inference}} \\ \cmidrule(lr){2-3} \cmidrule(lr){4-5}
 & \textbf{Peak Memory} & \textbf{Time-Consuming} & \textbf{Peak Memory} & \textbf{Lantency per Token} \\ \midrule
Fine-Tuning & 37.43 GB & 53 h & 17.71 GB & 34 ms \\ \midrule
Replay (1\%) & 37.43 GB & \textcolor[HTML]{ff644e}{54 h} & 17.71 GB & 34 ms \\
Replay (10\%) & 37.43 GB & \textcolor[HTML]{ff644e}{59 h} & 17.71 GB & 34 ms \\
EWC & \textcolor[HTML]{ff644e}{38.73 GB} & \textcolor[HTML]{ff644e}{54 h} & 17.71 GB & 34 ms \\
PathWeave & \textcolor[HTML]{ff644e}{40.08 GB} & \textcolor[HTML]{ff644e}{81 h} & \textcolor[HTML]{ff644e}{20.32 GB} & \textcolor[HTML]{ff644e}{111 ms} \\ \midrule
MERA (1\%) & 37.43 GB & \textcolor[HTML]{ff644e}{54 h} & 17.71 GB & 34 ms \\
MERA (10\%) & 37.43 GB & \textcolor[HTML]{ff644e}{61 h} & 17.71 GB & 34 ms \\ \bottomrule
\end{tabular}
\caption{Training and inference overheads of different methods. The peak memories during training and inference are measured with batch sizes of 4 and 1 respectively. The time-consuming refers to the total GPU hours for continually learning the four modalities. All metrics are measured on a single NVIDIA RTX A6000 48G. The non-optimal results are colored in \textcolor[HTML]{ff644e}{red}.}
\label{tab:efficiency_comparison}
\end{table*}

\begin{table}[!t]
\centering
\resizebox{\columnwidth}{!}{
\begin{tabular}{lcccc}
\toprule
\multirow{2}{*}{\textbf{Method}} & \multicolumn{2}{c}{\textbf{Sequential}} & \multicolumn{2}{c}{\textbf{Reverse}} \\ \cmidrule(lr){2-3} \cmidrule(lr){4-5}
 & \textbf{Mean} & \textbf{Std} & \textbf{Mean} & \textbf{Std} \\ \midrule
MERA & \textbf{97.90} & 6.02 & \textbf{84.42} & \textbf{12.93} \\ \midrule
\begin{tabular}[c]{@{}l@{}}MERA\\ w/ 10\% noise\end{tabular} & 96.87 & \textbf{4.97} & 83.56 & 13.48 \\ \midrule
\begin{tabular}[c]{@{}l@{}}MERA\\ w/ 50\% noise\end{tabular} & 94.10 & 5.67 & 79.79 & 15.54 \\ \bottomrule
\end{tabular}
}
\caption{Evaluations of MERA's robustness to noisy replay data. The realigning stage uses 10\% replay data.}
\label{tab:noisy_data}
\end{table}

\subsection{Robustness to the Quality of Replay Data}
Since the realigning stage of MERA relies on a small replay dataset, it is beneficial to understand how robust MERA is to the quality of replay data.
To evaluate its robustness, we manually corrupt $p\%$ samples in the replay dataset by mispairing their text-modality pair, and test MERA's performance.
Results in Table~\ref{tab:noisy_data} suggest that 10\% noisy samples does not significantly corrupt MERA's performance, and it still shows decent performance even under the extreme condition of 50\% noisy samples.
This suggests that our MERA is robust to low-quality replay data.

\subsection{Why is the Performance Degradation More Severe in Reverse Training Order}
In Table~\ref{tab:bw_relative_gain}, it is observed that the performance degradation is more severe in reverse training order than in sequential training order.
This phenomenon occurs across different methods.
We conjecture that training in sequential order is a sort of curriculum learning, while the reverse order corresponds to reversed curriculum learning.
In our experiments, the sequential order is determined based on the prevalence of different modalities, and more prevalent modalities might be easier to learn.
For example, the image modality is easier to understand than video, and the point cloud might be the most difficult modality to learn.
\citet{inscl} also observed that the performance degradation is less severe when continually learning in a curriculum learning order than in a reversed curriculum learning order.
We conjecture that learning easy knowledge leads to the forgetting of harder ones, but learning hard knowledge might even consolidate the easier ones.

\begin{table}[!t]
\centering
\resizebox{\columnwidth}{!}{
\begin{tabular}{cccc}
\toprule
 & \#Training Set & \#Test Set & License \\ \midrule
MSCOCO-2014* & 82K & 1K & CC-BY 4.0 \\
OK-VQA & 26K & 1K & CC-BY 4.0 \\
MSVD & 48K & 670 & - \\
MSVD-QA & 30K & 1K & - \\
AudioCaps & 44K & 1K & - \\
Clotho-AQA* & 15K & 1K & MIT License \\
Cap3D* & 50K & 1K & ODC-BY 1.0 \\
Cap3D-QA* & 30K & 1K & - \\ \bottomrule
\end{tabular}
}
\caption{Statistics of the datasets. Datasets marked with * are filtered from their original ones.}
\label{tab:dataset}
\end{table}

\begin{table*}[!t]
\centering
\begin{tabular}{cccc}
\toprule
Hyperparameters & Pre-Training & Fine-Tuning & Realigning \\ \midrule
Trainable Components & Connectors & LLM and Connectors & Connectors \\
Batch Size & 128 & 16 & 16 \\
Learning Rate of Connectors & 1e-3 & 2e-5 & 2e-5 \\
Learning Rate of LLM & - & 2e-4 & - \\
Learning Rate Schedule & \multicolumn{3}{c}{Cosine Decay} \\
Warmup Ratio & \multicolumn{3}{c}{0.03} \\
Epoch & \multicolumn{3}{c}{1} \\ \bottomrule
\end{tabular}
\caption{Hyperparameters for each training process. Pre-Training and Fine-Tuning refer to the two stages of the standard MLLM training process.}
\label{tab:hyperparameters}
\end{table*}

\section{Dataset Details}\label{appendix:datasets}
Table~\ref{tab:dataset} details the statistics of each dataset.
Some datasets are filtered from their original ones:
\begin{itemize}
\item MSCOCO-2014 \citep{mscoco}: Each image has multiple captions, we only use its first caption to form the training set.
\item Clotho-AQA \citep{clotho_aqa}: Each sample is annotated with a confidence level, we only use the samples whose confidence levels are "yes" to form the training set and test set.
\item Cap3D \citep{cap3d}: Since the original dataset is huge in scale, we filter out the samples whose caption is longer than 100 letters. Then, we randomly sample a 50K subset as the training set.
\item Cap3D-QA \citep{x_instructblip}: Since the original dataset is huge in scale, we randomly sample a 30K subset as the training set.
\end{itemize}
For each dataset, we use a randomly sampled 1K subset of its holdout test set as the final test set, except for the MSVD \citep{msvd}, since the size of its original test set is less than 1K.

\section{Experimental Details}

\subsection{Implementation Details}\label{appendix:implementation}
We build our experimental codebase on top of LLaVA \citep{llava, llava_1.5} and NExT-GPT \citep{next_gpt}.
We detail the modality-specific components of each modality as follows:
\begin{itemize}
\item Image: We use CLIP-ViT-L-336px \citep{clip_vit_l_336px} as the pre-trained image encoder, a randomly initialized MLP as the connector.
\item Video: We use CLIP-ViT-L-336px \citep{clip_vit_l_336px} as the pre-trained video encoder, a randomly initialized MLP as the connector. We uniformly sample 4 frames from a video as input frames. Then each frame is encoded by the video encoder separately. The output feature frames are downsampled by 2x using bilinear pooling before sending into the connector to improve efficiency.
\item Audio: We use $\mathrm{BEATs_{iter3+}(AS2M)}$ \citep{beats} as the pre-trained audio encoder, a Q-Former \citep{blip2} initialized from the pre-trained bert-base-uncased \citep{bert} as the connector. The number of query tokens is set to 32.
\item Point Cloud: We use Point-BERT-v1.2 \citep{pointllm} as the pre-trained point cloud encoder, a randomly initialized MLP as the connector.
\end{itemize}
We set the hyperparameters mainly following previous works \citep{llava, llava_1.5}, as listed in Table~\ref{tab:hyperparameters}.
For training that involves updating the LLM backbone, we utilize parameter-efficient fine-tuning with LoRA \citep{lora} applied across all linear modules within the LLM, setting the LoRA rank to 128 and the alpha parameter to 128.
All the experiments are conducted on a single NVIDIA RTX A6000 48G with FP16.

\begin{table}[!t]
\centering
\resizebox{\columnwidth}{!}{
\begin{tabular}{|p{8.2cm}|}
\hline
\textbf{System Prompt:} \\ \hline
You are an intelligent chatbot designed for evaluating the correctness of generative outputs for question-answer pairs. Your task is to compare the predicted answer with the correct answer and determine if they match meaningfully. Here's how you can accomplish the task: \\
\#\#INSTRUCTIONS: \\
- Focus on the meaningful match between the predicted answer and the correct answer. \\
- Consider synonyms or paraphrases as valid matches. \\
- Evaluate the correctness of the prediction compared to the answer. \\ \hline
\textbf{User Prompt:} \\ \hline
Please evaluate the following question-answer pair: \\
 \\
Question: \colorbox[HTML]{ff968d}{<question>} \\
Correct Answer: \colorbox[HTML]{ff968d}{<answer>} \\
Predicted Answer: \colorbox[HTML]{ff968d}{<prediction>} \\
 \\
Provide your evaluation only as a yes/no and score where the score is an integer value between 0 and 5, with 5 indicating the highest meaningful match. Please generate the response in the form of a Python dictionary string with keys 'pred' and 'score', where value of 'pred' is a string of 'yes' or 'no' and value of 'score' is in INTEGER, not STRING.DO NOT PROVIDE ANY OTHER OUTPUT TEXT OR EXPLANATION. Only provide the Python dictionary string. For example, your response should look like this: \{'pred': 'yes', 'score': 4.8\}. \\ \hline
\end{tabular}
}
\caption{Prompts to query the GPT for open-ended QA evaluation. The placeholders in \colorbox[HTML]{ff968d}{red boxes} are filled according to each evaluated sample.}
\label{tab:gpt_judge_template}
\end{table}

\subsection{Evaluation Details}
For the calculation of CIDEr scores \citep{cider}, we utilized an open-sourced library CaptionMetrics \citep{cider_lib}.
For the calculation of prediction accuracy, we leverage a GPT-based open-ended QA evaluation with GPT-4o mini as the judge model. The GPT is prompted to judge whether the generated prediction semantically matches the ground truth answer. The prompt template is shown in Table~\ref{tab:gpt_judge_template}.
All the reported experimental results are from single runs.

\subsection{Implementation of Baselines}\label{appendix:baselines}
The implementation details of each CL baseline are listed as follows:
\begin{itemize}
\item \textbf{Replay} leverage a small replay dataset $R_i\leftarrow$ randomly sample $r$\% data from $\{D_1, D_2, \dots, D_{i-1}\}$. When training on a new modality $M_i$, it trains on the joint dataset of $R_i$ and $D_i$.
\item \textbf{EWC} firstly estimates the Fisher information matrix $F_{i-1}$ of the last training stage $i-1$ as:
\begin{equation*}
\begin{aligned}
&F_{i-1}=\\
&\mathbb{E}_{x\sim D_{i-1}}\nabla_{\theta_{i-1}}\mathcal{L}(\theta_{i-1}, x)\cdot \nabla_{\theta_{i-1}}\mathcal{L}(\theta_{i-1}, x)^T
\end{aligned}
\end{equation*}
where $\mathcal{L}(\theta_{i-1}, x)$ denotes the auto-regressive loss of model $\theta_{i-1}$ on data $x\sim D_{i-1}$, which is sampled from a 1\% size random subset of $D_{i-1}$.
Then the loss function $\mathcal{L}^*(\theta_i, x)$ of stage $i$ is:
\begin{equation*}
\mathcal{L}^*(\theta_i, x)=\mathcal{L}(\theta_i, x) + \sum_{j=0}^{i-1}\frac{\lambda}{2} F_j (\theta_i - \theta_{i-1})^2
\end{equation*}
where the hyperparameter $\lambda$ is set to 1 as default.
This implementation of $\mathcal{L}^*(\theta_i, x)$ is known as Online EWC \citep{online_ewc_1, online_ewc_2}.
\item \textbf{PathWeave} leverages Adapter-in-Adapter (AnA) modules. In our implementation, the AnA modules are injected into the LLM rather than the connector for a fair comparison. The rank of AnA is consistent with the LoRA rank of other baselines, which is 128. In their original settings, PathWeave removes the newly added modules when testing the former modalities. However, this results in the inability to perform cross-modality tasks, which are common in real applications. Therefore, we do not remove them for a fair comparison.
\end{itemize}

For each baseline, all the common parameters about training MLLM itself are the same and set as default in Table~\ref{tab:hyperparameters}.

\section{Complete Raw Data}\label{appendix:raw_data}
Table~\ref{tab:raw_data_sequential} and Table~\ref{tab:raw_data_reverse} show the raw data of Figure~\ref{fig:bw_relative_gain} and Figure~\ref{fig:fw_relative_gain}.

\begin{table*}[!t]
\centering
\resizebox{0.85\textwidth}{!}{
\begin{tabular}{llccccccccl}
\cline{1-10}
\multirow{2}{*}{Method} & \multirow{2}{*}{} & \multicolumn{2}{c}{Image} & \multicolumn{2}{c}{Video} & \multicolumn{2}{c}{Audio} & \multicolumn{2}{c}{Point Cloud} &  \\ \cline{3-10}
 &  & MSCOCO & OK-VQA & MSVD & MSVD-QA & AudioCaps & Clotho-AQA & Cap3D & Cap3D-QA &  \\ \cline{1-10}
\multicolumn{2}{l}{Individually Trained Experts} & 100.76 & 0.358 & 138.39 & 0.460 & 60.14 & 0.658 & 99.93 & 0.568 &  \\ \cline{1-10}
\multirow{4}{*}{Fine-Tuning} & Stage 1 & 100.76 & 0.358 & - & - & - & - & - & - &  \\
 & Stage 2 & 54.52 & 0.172 & 130.22 & 0.555 & - & - & - & - &  \\
 & Stage 3 & 34.87 & \textcolor[HTML]{72b85d}{0.303} & 12.78 & 0.292 & 43.17 & 0.590 & - & - &  \\
 & Stage 4 & \textcolor[HTML]{72b85d}{58.63} & 0.201 & \textcolor[HTML]{72b85d}{29.55} & \textcolor[HTML]{72b85d}{0.350} & 8.28 & 0.094 & 84.40 & 0.524 &  \\ \cline{1-10}
\multirow{4}{*}{Replay (1\%)} & Stage 1 & 100.76 & 0.358 & - & - & - & - & - & - &  \\
 & Stage 2 & 41.45 & 0.125 & 137.07 & 0.569 & - & - & - & - &  \\
 & Stage 3 & \textcolor[HTML]{72b85d}{65.79} & \textcolor[HTML]{72b85d}{0.276} & 30.74 & 0.312 & 55.21 & 0.675 & - & - &  \\
 & Stage 4 & 59.94 & 0.225 & \textcolor[HTML]{72b85d}{102.16} & \textcolor[HTML]{72b85d}{0.469} & 22.17 & 0.490 & 81.43 & 0.508 &  \\ \cline{1-10}
\multirow{4}{*}{Replay (10\%)} & Stage 1 & 100.76 & 0.358 & - & - & - & - & - & - &  \\
 & Stage 2 & 50.65 & 0.266 & 137.67 & 0.584 & - & - & - & - &  \\
 & Stage 3 & \textcolor[HTML]{72b85d}{83.32} & \textcolor[HTML]{72b85d}{0.318} & 33.87 & 0.381 & 44.82 & 0.651 & - & - &  \\
 & Stage 4 & 67.42 & 0.259 & \textcolor[HTML]{72b85d}{133.43} & \textcolor[HTML]{72b85d}{0.520} & 24.13 & 0.525 & 73.19 & 0.515 &  \\ \cline{1-10}
\multirow{4}{*}{EWC} & Stage 1 & 100.76 & 0.358 & - & - & - & - & - & - &  \\
 & Stage 2 & 64.84 & 0.208 & 155.09 & 0.595 & - & - & - & - &  \\
 & Stage 3 & 44.22 & \textcolor[HTML]{72b85d}{0.211} & 73.14 & 0.569 & 59.86 & 0.690 & - & - &  \\
 & Stage 4 & \textcolor[HTML]{72b85d}{56.54} & \textcolor[HTML]{72b85d}{0.227} & 36.72 & 0.564 & 26.64 & 0.651 & 96.40 & 0.551 &  \\ \cline{1-10}
\multirow{4}{*}{EProj} & Stage 1 & 92.16 & 0.298 & - & - & - & - & - & - &  \\
 & Stage 2 & 92.16 & 0.298 & 123.04 & 0.470 & - & - & - & - & \multicolumn{1}{c}{} \\
 & Stage 3 & 92.16 & 0.298 & 123.04 & 0.470 & 54.85 & 0.637 & - & - & \multicolumn{1}{c}{} \\
 & Stage 4 & 92.16 & 0.298 & 123.04 & 0.470 & 54.85 & 0.637 & 58.59 & 0.349 & \multicolumn{1}{c}{} \\ \cline{1-10}
\multirow{4}{*}{PathWeave} & Stage 1 & 100.76 & 0.358 & - & - & - & - & - & - &  \\
 & Stage 2 & 78.06 & 0.234 & 158.51 & 0.606 & - & - & - & - &  \\
 & Stage 3 & \textcolor[HTML]{72b85d}{79.07} & \textcolor[HTML]{72b85d}{0.251} & 138.63 & 0.547 & 59.47 & 0.682 & - & - &  \\
 & Stage 4 & 66.92 & \textcolor[HTML]{72b85d}{0.255} & 123.39 & 0.536 & 38.53 & 0.639 & 97.32 & 0.554 &  \\ \cline{1-10}
\multirow{4}{*}{MERA (1\%)} & Stage 1 & 100.76 & 0.358 & - & - & - & - & - & - &  \\
 & Stage 2 & 93.70 & 0.304 & 153.73 & 0.573 & - & - & - & - &  \\
 & Stage 3 & 90.42 & \textcolor[HTML]{72b85d}{0.316} & 147.42 & 0.567 & 57.09 & 0.678 & - & - &  \\
 & Stage 4 & \textcolor[HTML]{72b85d}{95.18} & \textcolor[HTML]{72b85d}{0.334} & 142.67 & 0.562 & 53.04 & 0.678 & 79.32 & 0.454 &  \\ \cline{1-10}
\multirow{4}{*}{MERA (10\%)} & Stage 1 & 100.76 & 0.358 & - & - & - & - & - & - &  \\
 & Stage 2 & 98.30 & 0.340 & 152.20 & 0.579 & - & - & - & - &  \\
 & Stage 3 & 96.46 & \textcolor[HTML]{72b85d}{0.346} & 147.89 & 0.566 & 61.49 & 0.684 & - & - &  \\
 & Stage 4 & \textcolor[HTML]{72b85d}{98.05} & 0.338 & 141.25 & 0.560 & 56.79 & 0.678 & 87.59 & 0.468 &  \\ \cline{1-10}
\end{tabular}
}
\caption{Raw data of sequential order training. Results that are better than the last stage are colored in \textcolor[HTML]{72b85d}{green}, indicating a Positive Backward Transfer.}
\label{tab:raw_data_sequential}
\end{table*}

\begin{table*}[!t]
\centering
\resizebox{0.85\textwidth}{!}{
\begin{tabular}{llcccccccc}
\hline
\multirow{2}{*}{Method} & \multirow{2}{*}{} & \multicolumn{2}{c}{Point Cloud} & \multicolumn{2}{c}{Audio} & \multicolumn{2}{c}{Video} & \multicolumn{2}{c}{Image} \\ \cline{3-10} 
 &  & Cap3D & Cap3D-QA & AudioCaps & Clotho-AQA & MSVD & MSVD-QA & MSCOCO & OK-VQA \\ \hline
\multicolumn{2}{l}{Individually Trained Experts} & 99.93 & 0.568 & 60.14 & 0.658 & 138.39 & 0.460 & 100.76 & 0.358 \\ \hline
\multirow{4}{*}{Fine-Tuning} & Stage 1 & 99.93 & 0.568 & - & - & - & - & - & - \\
 & Stage 2 & 2.74 & 0.178 & 39.25 & 0.519 & - & - & - & - \\
 & Stage 3 & \textcolor[HTML]{72b85d}{37.26} & \textcolor[HTML]{72b85d}{0.280} & 21.34 & 0.158 & 121.29 & 0.550 & - & - \\
 & Stage 4 & 26.69 & 0.199 & \textcolor[HTML]{72b85d}{23.11} & \textcolor[HTML]{72b85d}{0.519} & 23.40 & 0.266 & 86.12 & 0.342 \\ \hline
\multirow{4}{*}{Replay (1\%)} & Stage 1 & 99.93 & 0.568 & - & - & - & - & - & - \\
 & Stage 2 & 0.98 & 0.101 & 48.48 & 0.640 & - & - & - & - \\
 & Stage 3 & \textcolor[HTML]{72b85d}{35.18} & \textcolor[HTML]{72b85d}{0.337} & 8.30 & 0.138 & 124.89 & 0.546 & - & - \\
 & Stage 4 & 9.39 & 0.192 & \textcolor[HTML]{72b85d}{17.67} & \textcolor[HTML]{72b85d}{0.498} & 1.06 & 0.255 & 83.38 & 0.347 \\ \hline
\multirow{4}{*}{Replay (10\%)} & Stage 1 & 99.93 & 0.568 & - & - & - & - & - & - \\
 & Stage 2 & 0.63 & 0.171 & 47.38 & 0.641 & - & - & - & - \\
 & Stage 3 & \textcolor[HTML]{72b85d}{68.26} & \textcolor[HTML]{72b85d}{0.470} & 12.83 & 0.372 & 134.07 & 0.575 & - & - \\
 & Stage 4 & 3.42 & 0.241 & \textcolor[HTML]{72b85d}{18.33} & \textcolor[HTML]{72b85d}{0.540} & 2.81 & 0.228 & 87.31 & 0.342 \\ \hline
\multirow{4}{*}{EWC} & Stage 1 & 99.93 & 0.568 & - & - & - & - & - & - \\
 & Stage 2 & 29.97 & 0.442 & 59.31 & 0.672 & - & - & - & - \\
 & Stage 3 & 19.91 & 0.375 & 29.25 & 0.611 & 148.26 & 0.578 & - & - \\
 & Stage 4 & \textcolor[HTML]{72b85d}{45.25} & 0.327 & 23.39 & 0.511 & 47.53 & 0.524 & 98.91 & 0.320 \\ \hline
\multirow{4}{*}{EProj} & Stage 1 & 58.59 & 0.349 & - & - & - & - & - & - \\
 & Stage 2 & 58.59 & 0.349 & 54.85 & 0.637 & - & - & - & - \\
 & Stage 3 & 58.59 & 0.349 & 54.85 & 0.637 & 123.04 & 0.470 & - & - \\
 & Stage 4 & 58.59 & 0.349 & 54.85 & 0.637 & 123.04 & 0.470 & 92.16 & 0.298 \\ \hline
\multirow{4}{*}{PathWeave} & Stage 1 & 99.93 & 0.568 & - & - & - & - & - & - \\
 & Stage 2 & 71.79 & 0.420 & 55.07 & 0.648 & - & - & - & - \\
 & Stage 3 & 63.75 & 0.380 & 38.73 & 0.628 & 148.61 & 0.577 & - & - \\
 & Stage 4 & 55.23 & 0.370 & 37.23 & 0.603 & 85.67 & 0.521 & 87.04 & 0.361 \\ \hline
\multirow{4}{*}{MERA (1\%)} & Stage 1 & 99.93 & 0.568 & - & - & - & - & - & - \\
 & Stage 2 & 64.79 & 0.470 & 58.77 & 0.684 & - & - & - & - \\
 & Stage 3 & \textcolor[HTML]{72b85d}{66.76} & 0.377 & 43.00 & 0.595 & 147.99 & 0.547 & - & - \\
 & Stage 4 & \textcolor[HTML]{72b85d}{70.40} & \textcolor[HTML]{72b85d}{0.387} & \textcolor[HTML]{72b85d}{51.02} & \textcolor[HTML]{72b85d}{0.651} & 140.20 & 0.538 & 93.33 & 0.362 \\ \hline
\multirow{4}{*}{MERA (10\%)} & Stage 1 & 99.93 & 0.568 & - & - & - & - & - & - \\
 & Stage 2 & 87.10 & 0.505 & 58.99 & 0.695 & - & - & - & - \\
 & Stage 3 & 79.24 & 0.437 & 58.65 & 0.650 & 145.56 & 0.552 & - & - \\
 & Stage 4 & \textcolor[HTML]{72b85d}{81.05} & 0.425 & \textcolor[HTML]{72b85d}{60.28} & \textcolor[HTML]{72b85d}{0.653} & \textcolor[HTML]{72b85d}{146.72} & \textcolor[HTML]{72b85d}{0.569} & 97.62 & 0.367 \\ \hline
\end{tabular}
}
\caption{Raw data of reverse order training. Results that are better than the last stage are colored in \textcolor[HTML]{72b85d}{green}, indicating a Positive Backward Transfer.}
\label{tab:raw_data_reverse}
\end{table*}

\end{document}